\def\year{2020}\relax
\documentclass[letterpaper]{article} 
\usepackage[utf8]{inputenc}
\usepackage[T1]{fontenc}
\usepackage{aaai20}  
\usepackage{times}  
\usepackage{helvet} 
\usepackage{courier}  
\usepackage[hyphens]{url}  
\usepackage{graphicx} 
\usepackage{graphicx}  
\nocopyright
\usepackage{amsmath}
\usepackage[capitalize]{cleveref}

\title{Towards Interpretable Semantic Segmentation via \\ 
Gradient-weighted Class Activation Mapping }
\author{%
Kira Vinogradova,
Alexandr Dibrov,
Gene Myers
\\
Max Planck Institute of Molecular Cell Biology and Genetics, Dresden, Germany\\
Center for Systems Biology Dresden, Germany\\[1mm]
\texttt{vinograd@mpi-cbg.de}
}

\newcommand{\sgc}{\mbox{\small\textsc{seg-grad-cam}}}

\begin{document}

\maketitle

\begin{abstract}
Convolutional neural networks have become state-of-the-art in a wide range of image recognition tasks. The interpretation of their predictions, however, is an active area of research. Whereas various interpretation methods have been suggested for image classification, the interpretation of image segmentation still remains largely unexplored. To that end, we propose \sgc, a gradient-based method for interpreting semantic segmentation. Our method is an extension of the widely-used Grad-CAM method, applied locally to produce heatmaps showing the relevance of individual pixels for semantic segmentation.
\end{abstract}

\section{Introduction}
\label{sec:intro}

Approaches based on deep learning, and convolutional neural networks (CNNs) in particular, have recently substantially improved the performance for various image understanding tasks, such as image classification, object detection, and image segmentation.
However, our understanding of \emph{why} and \emph{how} CNNs achieve state-of-the-art results is rather immature.

One avenue to remedy this is to visually indicate which regions of an input image are (especially) important
for the decision made by a CNN.
These so-called \emph{heatmaps} can thus be useful to understand a CNN, for example to check that it does not focus on idiosyncratic details of the training images that will not generalize to unseen images.

Gradient-based heatmap methods have generally been popular in the context of image classification.
A simple approach are \emph{saliency maps}~\cite{simonyan2013deep}, which are obtained via the derivative of the logit $y^c$ (the score of class $c$ before the softmax) with respect to all pixels of the input image.
Hence, they highlight pixels whose change would affect the score of class $c$ the most.
A more recent and widely-used method by \citeauthor{selvaraju2017grad} \citeyear{selvaraju2017grad} is \emph{gradient-weighted class activation mapping} (Grad-CAM).
It first uses the aggregated gradients of logit $y^c$ with respect to chosen feature layers to determine their general relevance for the decision of the network. Based on this relevance, a heatmap is obtained as a weighted average of the activations of the respective feature layers (feature maps).
Grad-CAM can be seen as a generalization of CAM \cite{zhou2016learning}
, which could only produce class activation mappings for CNNs with a special architecture.

\begin{figure}[t]
\centering
\includegraphics[width=1\linewidth,
]{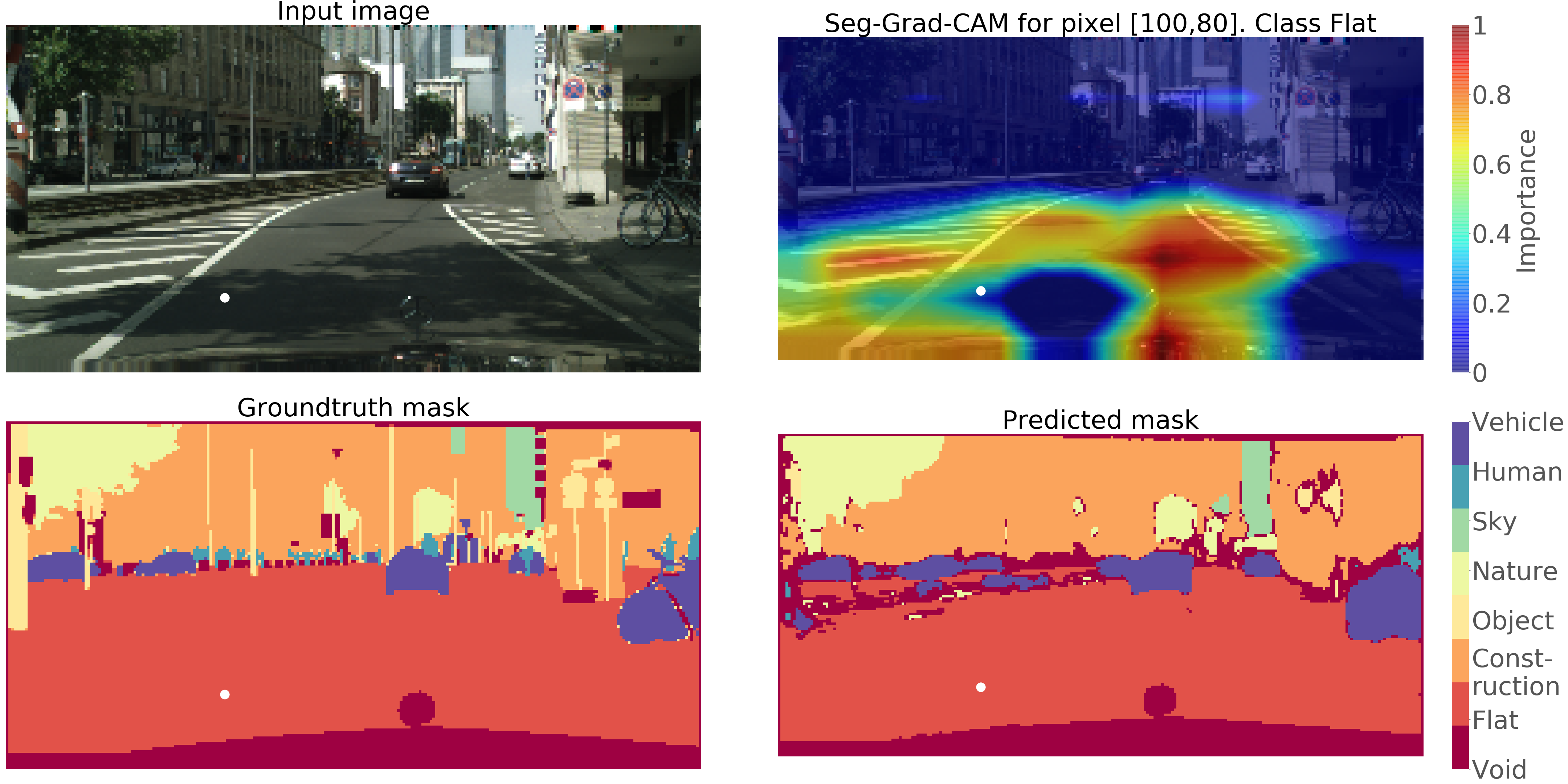}%
\caption{%
\sgc\ for a single pixel (white dot) and class \emph{Flat}.
The heatmap is obtained with
respect to a convolutional layer at the bottleneck (i.e.~end of contracting path) of a U-Net~\cite{ronneberger2015u}.
}
\label{fig:example_cityscape}
\end{figure}

Methods that provide visual explanations for the decisions of neural networks have predominantly focused on the task of image classification.
In this work, we go beyond that and are interested in explaining the decisions of CNNs for semantic image segmentation.
To that end, we propose \sgc, an extension of Grad-CAM for semantic segmentation, which can produce heatmaps that explain the relevance for the decision of individual pixels or regions in the input image.
We demonstrate that our approach produces reasonable visual explanations for the commonly-used
Cityscapes datasets \cite{Cordts2016Cityscapes}.

Concurrent to our work, \citeauthor{hoyer2019grid} have independently proposed a method for the visual explanation of semantic segmentation CNNs \cite{hoyer2019grid}. They assume co-occurences of some classes are important for their segmentation. However, their approach is not based on Grad-CAM, but on perturbation analysis, and is rather different from ours since it focuses on identification of contextual biases.

To the best of our knowledge, we present the first approach to produce visual explanations of CNNs for semantic segmentation, specifically by extending Grad-CAM.

\section{Method}
\label{sec:method}

As mentioned above, our approach is based on Grad-CAM~\cite{selvaraju2017grad}, which we first briefly explain.
Let $\{A^k\}_{k=1}^K$ be selected feature maps of interest ($K$ kernels of the last convolutional layer of a classification network), and $y^c$ the logit for a chosen class $c$.
Grad-CAM averages the gradients of $y^c$ with respect to all $N$ pixels (indexed by $u,v$) of each feature map $A^k$
to produce a weight $\alpha_c^k$ to denote its importance.
The heatmap
\begin{equation}
    \label{eq:grad-cam}
    L^c = \mathrm{ReLU}\biggl(\sum_k \alpha_c^k A^k \biggr)
    \ \ \text{with}\ \
    \alpha_c^k = \frac{1}{N} \sum_{u,v} \frac{\partial y^c}{\partial A_{uv}^k}
\end{equation}
is then generated by using these weights to sum the feature maps;
finally, $\mathrm{ReLU}$ is applied pixel-wise to clip negative values at zero,
to only highlight areas that positively contribute to the decision for class $c$.

Whereas a classification network predicts a single class distribution
per input image $x$, a CNN for semantic segmentation typically
produces logits $y_{ij}^c$ for every pixel $x_{ij}$ and class $c$.

Hence, we propose \sgc\  by replacing
$y^c$ by $\sum_{(i,j) \in \mathcal{M}} y_{ij}^c$ in \cref{eq:grad-cam},
where $\mathcal{M}$ is a set of pixel indices of interest in the output mask. 
This allows to adapt Grad-CAM to a semantic segmentation network
in a flexible way, since $\mathcal{M}$ can denote just a single pixel,
or pixels of an object instance, or simply all pixels of the image.
Furthermore, we explore using feature maps from
intermediate convolutional layers, not only the last one as used in
\citeauthor{selvaraju2017grad} \citeyear{selvaraju2017grad}.

\section{Experiments}
\label{sec:experiments}

We demonstrate our approach by training a U-Net~\cite{ronneberger2015u} for semantic segmentation
of the popular \emph{Cityscapes} dataset \cite{Cordts2016Cityscapes}.
We generally find that the convolutional layers of the U-Net bottleneck (end of the encoder before upsampling)
are more informative than the layers close to the end of the U-Net decoder, which would be more similar
to those inspected by \citeauthor{selvaraju2017grad} \citeyear{selvaraju2017grad}.
As a sanity check, we do observe (not shown) that heatmaps produced from the initial convolutional layers
exhibit edge-like structures, which does agree with common knowledge that early
convolutional layers pick up on low-level image features.
Feature maps located between the bottleneck and last layer successively give rise to heatmaps that look more and more similar to the logits of the selected class and the output segmentation mask.

\begin{figure}[t]
\centering
\includegraphics[width=1\linewidth,%
]{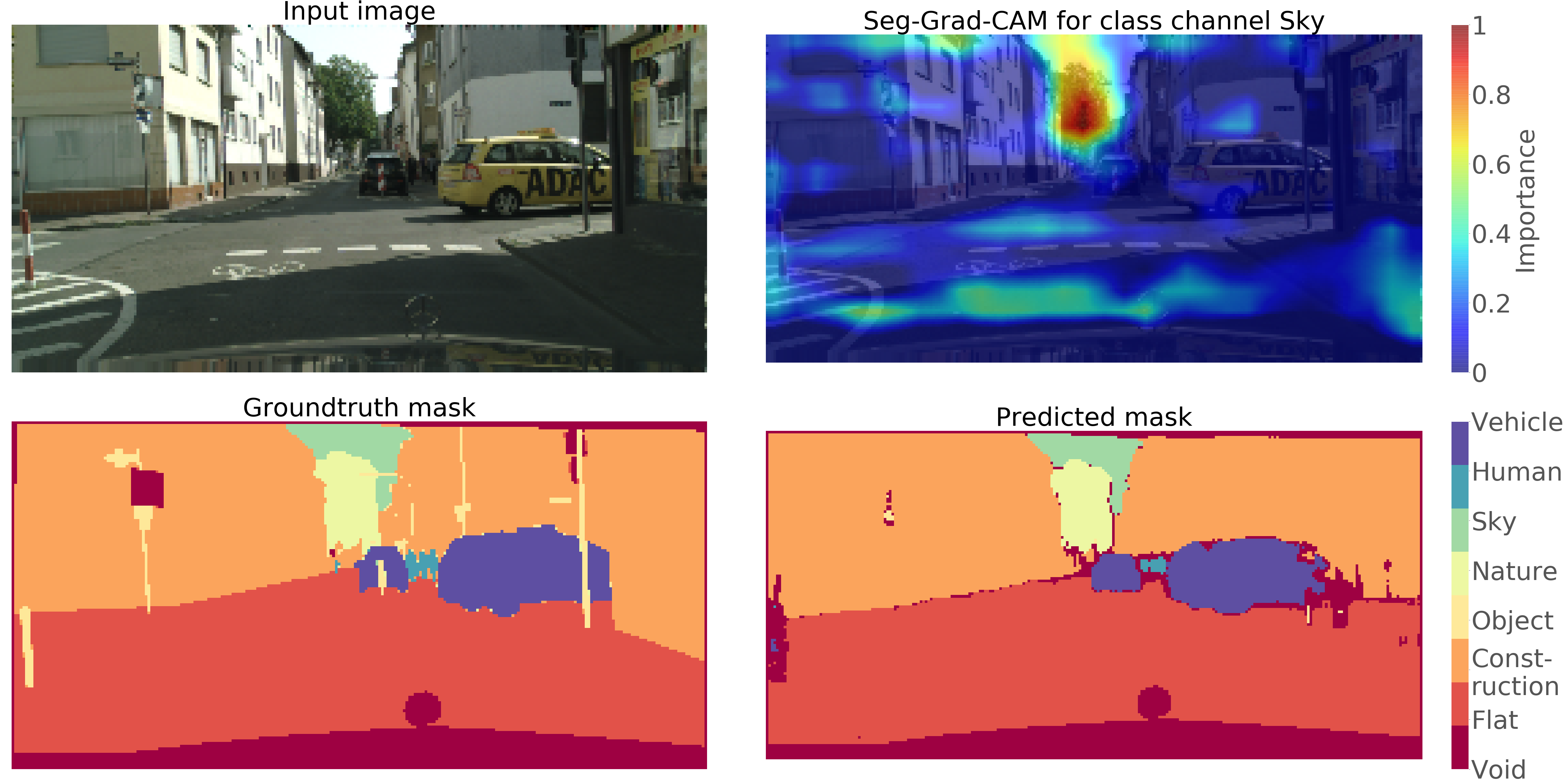}%
\caption{%
\sgc\ for all pixels and class \emph{Sky}.
The heatmap is obtained with
respect to a convolutional layer at the bottleneck (i.e.~end of contracting path) of a U-Net~\cite{ronneberger2015u}.
}
\label{fig:example_cityscape2}
\end{figure}

\cref{fig:example_cityscape} shows a heatmap produced by \sgc\ for a
bottleneck layer of the U-Net when $\mathcal{M}$ denotes a single pixel. The
visually highlighted region seems plausible, mostly indicating similar pixels of
the selected class.
Note that the heatmap shows the weighted sum of feature maps activated for the whole
image (cf.~Eq.~\ref{eq:grad-cam}), and can thus go beyond the receptive field of the CNN for the selected pixel,
whose relevance is only for determining the weights $\alpha_c^k$.
Furthermore, \cref{fig:example_cityscape2} shows a heatmap for class \emph{Sky}
when $\mathcal{M}$ indicates all pixels of the image; it most strongly highlights pixels of a
tree (class \emph{Nature}), which may be highly informative to predict \emph{Sky} pixels.

\section{Discussion and Future Work}
\label{sec:discussion}

Our initial results seem promising, and we would like to systematically
investigate the generated heatmaps of our \sgc\ method in the future.
Concretely, we want to compare and reason about different intermediate feature
maps that can be chosen for visualization.
Furthermore, it might be helpful to truncate the extent of the heatmap only
to regions that are directly relevant for the prediction at pixels contained in $\mathcal{M}$.
For a fixed class $c$, it would also be interesting to compare the weights $\{\alpha_c^k\}_{k=1}^K$
as obtained at different locations.
Finally, we aim to explore other interpretation approaches \cite{montavon2018methods} and plan to demonstrate the merits of our method quantitatively, based on a suitable synthetic dataset.

\end{document}